%% file: mainV2.tex
\documentclass[times, review, 10pt]{elsarticle}

\usepackage{amssymb}
\usepackage{amsmath}
\usepackage{booktabs}
\usepackage{url}
\usepackage{hyperref}

\begin{document}

\begin{frontmatter}



\title{GRIT-LP: Graph Transformer with Long-Range Skip Connection and Partitioned Spatial Graphs for Accurate Ice Layer Thickness Prediction} %% Article title


%% Author name
\author[1]{Zesheng Liu}
\author[1,2]{Maryam Rahnemoonfar}
%% Author name

\affiliation[1]{organization={Department of Computer Science and Engineering, Lehigh University},%Department and Organization
            addressline={113 Research Drive, Building C}, 
            city={Bethlehem},
            postcode={18015}, 
            state={PA},
            country={USA}}
\affiliation[2]{organization={Department of Civil and Environmental Engineering, Lehigh University},%Department and Organization
            addressline={19 Memorial Drive W.}, 
            city={Bethlehem},
            postcode={18015}, 
            state={PA},
            country={USA}}

% \affiliation{organization={},%Department and Organization

%% Abstract
\begin{abstract}
Graph transformers have demonstrated remarkable capability on complex spatio-temporal tasks, yet their depth is often limited by oversmoothing and weak long-range dependency modeling. To address these challenges, we introduce GRIT-LP, a graph transformer explicitly designed for polar ice-layer thickness estimation from polar radar imagery. Accurately estimating ice layer thickness is critical for understanding snow accumulation, reconstructing past climate patterns and reducing uncertainties in projections of future ice sheet evolution and sea level rise. GRIT-LP combines an inductive geometric graph learning framework with self-attention mechanism, and introduces two major innovations that jointly address challenges in modeling the spatio-temporal patterns of ice layers: a partitioned spatial graph construction strategy that forms overlapping, fully connected local neighborhoods to preserve spatial coherence and suppress noise from irrelevant long-range links, and a long-range skip connection mechanism within the transformer that improves information flow and mitigates oversmoothing in deeper attention layers. We conducted extensive experiments, demonstrating that GRIT-LP outperforms current state-of-the-art methods with a 24.92\% improvement in root mean squared error. These results highlight the effectiveness of graph transformers in modeling spatiotemporal patterns by capturing both localized structural features and long-range dependencies across internal ice layers, and demonstrate their potential to advance data-driven understanding of cryospheric processes.
\end{abstract}

% \includegraphics[width=\textwidth]{GRIT-LP-Graphical-Abstract.png}

% \begin{highlights}
% \item We propose GRIT-LP, a graph transformer for deep ice-layer thickness prediction.
% \item We design adaptive long-range skips from raw spatial embeddings to temporal blocks.
% \item We propose localized graph connectivity for graph construction.
% \item Extensive analysis shows GRIT-LP improves accuracy by 24.92\% over prior methods.
% \end{highlights}

%% Keywords

\end{frontmatter}



\input{introduction}

\input{relatedwork}

\input{keydesigns}

\input{dataset}
\input{experiments}

\input{results}

\input{conclusion}

\section*{Author contributions}

\textbf{Zesheng Liu:} Formal analysis, Investigation, Methodology, Software, Validation, Visualization, Writing-original draft, Writing-review\&editing

\textbf{Maryam Rahnemoonfar:} Conceptualization, Funding acquisition, Project administration, Resources, Supervision, Writing-review\&editing

\section*{Acknowledgement}

This work is supported by NSF BIGDATA awards (IIS-1838230, IIS-2308649), NSF Leadership Class Computing awards (OAC-2139536), NSF PFI awards (2423211). We acknowledge data and data products from CReSIS generated with support from the University of Kansas and NASA Operation Ice-Bridge.

% %% The Appendices part is started with the command \appendix;
% %% appendix sections are then done as normal sections
% \appendix
% \section{Example Appendix Section}


% %% For citations use: 
% %%       \citet{<label>} ==> Lamport [21]
% %%       \citep{<label>} ==> [21]
% %%
% Example citation, See \citet{lamport94}.

\bibliographystyle{elsarticle-num-names} 
\bibliography{ref}




% %% For authoryear reference style

% \bibitem[Lamport(1994)]{lamport94}
%   \textit{\LaTeX: a document preparation system},
%   Addison Wesley, Massachusetts,

% \end{thebibliography}
\end{document}

%% file: introduction.tex
\section{Introduction}

Graph transformers have proven to be highly effective for modeling complex graph-structured data, with wide-range of applications in real-world scenarios, particularly those involving spatiotemporal patterns. Their ability to capture intricate relationships and dependencies makes them highly valuable in domains such as pedestrian trajectory prediction \cite{YuMa_STGT_Pedestrian} and traffic prediction \cite{Intro_Traffic_Prediction}.

Despite their success, current graph transformer architectures face notable limitations, including overfitting and over-smoothing—a phenomenon where node features become indistinguishable as layers deepen \cite{chen2020oversmoothing}. Additionally, many existing graph transformers are relatively shallow, limiting their ability to effectively capture the complex, long-range dependencies that often emerge in real-world datasets. Addressing these challenges is essential, as deeper architectures are critical for modeling the hierarchical and intricate relationships necessary to understand spatiotemporal phenomena.

In this work, we address these limitations by introducing GRIT-LP, a graph transformer enhanced with novel long-range skip connections and partitioned spatial graphs. Specifically, GRIT-LP (\textbf{G}raph t\textbf{R}ansformer for \textbf{I}ce-layer \textbf{T}hickness with \textbf{L}ong-range skip connections and \textbf{P}artitioned spatial graphs) is designed to address the task of learning spatio-temporal patterns within ice layers, leveraging information from shallow internal layers to accurately predict the thickness of deeper layers. An accurate understanding of internal ice layer thickness and variability is essential for monitoring snow accumulation and assessing ice sheet dynamics. These insights not only improve climate models by reducing uncertainties but also help predict and mitigate the impacts of accelerating ice loss and rising global temperatures. Successfully modeling ice‐layer evolution requires representations that capture both the spatial coherence within each layer and the temporal progression across years. Graph‐based frameworks naturally encode layer interfaces and facilitate learning their inter‐layer relationships.

In order to learn the spatio-temporal patterns over ice layers formed in successive years, \citet{Zalatan2023,Zalatan_igarss, zalatan_icip} proposed to represent each individual ice layer as a graph, and developed a multi-target, recurrent graph convolutional neural network, AGCN-LSTM, to learn the relationship between shallow and deep ice layers. \citet{liu2024multibranchspatiotemporalgraphneural} proposed a novel multi-branch spatio-temporal graph neural network that decouples the learning process of spatial variations and temporal changes, which allows a better network weight optimization and improve the training efficiency. Considering the fact that graph convolutional networks are usually limited by their localized receptive fields and thus struggle to capture the long temporal dependencies, \citet{liu2025_GRIT,liu2025_STGRIT} also extended the network to graph transformers. By integrating self-attention into a geometric deep learning framework, these methods more effectively capture both local and global dependencies, leveraging hierarchical structure and long-range patterns that are critical for modeling temporal changes in ice-sheet behavior. However, in all these methods, graphs are fully connected, which imposes uniform interactions between all nodes regardless of their contextual relevance, making it difficult for the model to distinguish informative relationships from irrelevant ones and potentially diluting the learning of meaningful spatio-temporal dependencies.

To address challenges in current graph transformer networks and further improve the accuracy for ice layer thickness prediction, our proposed GRIT-LP integrates temporal attention into the geometric deep learning framework and incorporates long-range skip connections that dynamically balances raw spatial feature embeddings and transformed temporal features while preserving feature scale. We also introduce a novel way to define graph connectivity: locally fully connected within sliding spatial neighborhoods, while avoiding connections across distant regions. In this structure, nodes are grouped based on spatial proximity, and each group forms a fully connected subgraph. These groups slide across the spatial domain with fixed size and overlap, creating localized cliques that retain strong local interactions. This design enhances the model's ability to capture meaningful spatial patterns while maintaining sparsity in the graph structure, avoiding the pitfalls of indiscriminate global connectivity that may dilute informative relationships. The key contributions of this work are:

\begin{itemize}
    % \item To the best of our knowledge, we are the first to develop novel graph transformers on polar ice application, aiming to address the task of learning the spatio-temporal relationship between shallow and deep ice layers.
    
    \item We developed GRIT-LP, graph transformer for ice layer thickness with long-range skip connection and partitioned spatial graphs, that is designed to learn the patterns from the upper $l$ internal ice layers and predict the thickness for the underlying $m$ layers.
    
    \item Our proposed framework combines geometric deep learning for capturing spatial patterns within each layer as feature embeddings and temporal attention blocks to effectively learn both long-term and short-term temporal dependencies precisely.

    \item GRIT-LP framework introduces a novel adaptive long-range skip connection that balances raw spatial feature embeddings and transformed temporal features during the learning process, serving as a dynamic control that mitigates critical challenges in training such as overfitting and over-smoothing. This design allows the model to selectively preserve earlier representations and adaptively combine them with deeper features for richer and more stable representation.
    
    % , effectively handle deep architectures and reduce the uncertainty in model performance.
    % demonstrate that selectively preserving earlier representations and adaptively combining them with deeper features leads to richer and more stable representations, particularly when modeling hierarchical or long-range dependencies.

    \item We introduce a novel graph connectivity strategy based on locally fully connected spatial neighborhoods. Nodes are grouped into fixed-size, overlapping spatial windows, where each group forms a fully connected subgraph. This structure retains strong local spatial relationships while avoiding indiscriminate global connections, enabling more focused and efficient spatial representation learning within each ice layer.

    \item  We conduct extensive experiments comparing GRIT-LP against current state-of-the-art method and multiple recurrent graph convolutional network baselines, on a representative case where $l=5$ and $m=15$. Results shows that GRIT-LP achieves a 24.92\% improvement in root mean squared error compared with state-of-the-art method, highlighting its superior performance and robustness in ice layer thickness prediction tasks.
\end{itemize}

%% file: relatedwork.tex
\section{Related Work}

\subsection{Automatic Internal Ice Layer Tracking}
Identifying the boundaries of internal ice layers from the radargrams is a challenging task, as deep ice layers formed a long time ago may be broken or fully melted. Moreover, this task requires a scalable algorithm, as radar sensors usually collect massive data in a pretty short time. Deep learning approaches, including convolutional neural networks (CNNs) and generative adversarial networks (GANs), have been utilized to track ice layers from radargrams images\cite{DeepIceLayerTracking,DeepLearningOnAirborneRadar,Rahnemoonfar_2021_JOG,Yari_2021_JSTAR}. While these methods have reached some success, they frequently highlight persistent challenges, such as noise within input radargrams and the limited availability of high-quality snow radar datasets with reliable annotations. To address these challenges, recently some researchers tried to incorporate the idea of physics-informed learning, employing physics-based wavelet transform for denoising \cite{DeepHybridWavelet,varshney2021refining, Varshney_Yari_Ibikunle_Li_Paden_Gangopadhyay_Rahnemoonfar_2024} or generating high-quality labels from physical models to pretrain the networks \cite{LearnSnowLayerThickness}. 

Compared with these traditional methods, our proposed GRIT-LP represents internal ice layers as individual graph data and applies geometric deep learning and attention mechanisms, which are less sensitive to noise in the network input. This approach benefits from a more robust and effective learning process, ensuring a reliable performance across inputs with different qualities.

\subsection{Graph Transformers}
Graph transformer has been widely used in understanding complex patterns in real-world domain data, especially in spatial-temporal tasks like traffic prediction \cite{Dynamic_Graph_Transformer,STPT_Traffic,PR_TSANet}, human pose estimation\cite{PR_DGFormer}, molecular data processing and property estimation \cite{Chemical_GT_2, Gao2024_ChemicalGT}, and processing other chemical and biological data\cite{Do_Transformers_Really_Perform_Bad,mialon2021graphit,Gophormer}. Although these prior research has already demonstrated the outstanding performance of graph transformers, most studies primarily focus on refining the self-attention mechanism to better suit graph datasets and introducing effective feature encoding to preserve critical structural information. Little attention has been given to other building blocks of attention encoders. In contrast, our GRIT-LP network explores different types of skip connections and proposes a novel long-range skip connection that enables the attention encoder to go deeper, effectively capturing complex high-level patterns in real-world application data.

\subsection{Graph Neural Network and Graph Transformers For Ice Layer Thickness Prediction}
\citet{Zalatan2023,Zalatan_igarss, zalatan_icip} were the first to represent internal ice layers as independent graphs and applied graph neural networks to predict deep ice layer thickness. They introduced a multi-target adaptive model, AGCN-LSTM, which combines a graph convolutional network (GCN) with a long short-term memory (LSTM) network. By incorporating EvolveGCNH \cite{EGCN} as an adaptive layer, AGCN-LSTM effectively captures spatio-temporal patterns within internal ice layers.  \citet{liu2024multibranchspatiotemporalgraphneural} further improved the accuracy and efficiency of internal ice layer thickness estimation by applying a multi-branch structure that separates the learning process for spatial and temporal patterns. Later, \citet{liu2025_GRIT,liu2025_STGRIT} extends the networks architecture to graph transformers, where self-attention mechanism is combined with geometric deep learning framework. As the current state-of-the-art methods, their proposed ST-GRIT\cite{liu2025_STGRIT} contains GraphSAGE\cite{hamilton2018inductive_graphsage} layers to learn the spatial features from each input graph and combined together as feature embeddings. Transformer encoders are then applied to the feature embeddings on both the spatial and temporal dimension. 

Compared with current state-of-the-art methods, GRIT-LP removes explicit spatial attention blocks, allowing a deeper stack of temporal attention blocks at the same computation time. Instead of separate spatial attention, GRIT-LP introduces an adaptive long-range skip connection that links spatial feature embeddings to the output of each temporal attention block. This connection dynamically balances raw spatial embeddings and transformed temporal features while preserving feature scale, yielding a richer, more stable representation and enabling a deeper architecture to capture critical temporal relationships. Additionally, GRIT-LP applies a novel localized connectivity based on partitioned spatial graphs, which encourages structural sparsity and avoid irrelevant interactions among long-distance nodes.

%% file: keydesigns.tex
\section{Graph Transformer with Long-Range Skip Connection and Partitioned Spatial Graphs for Accurate Ice Layer Thickness Prediction}
\label{architecture}
\begin{figure*}[ht]
\begin{center}
\centerline{\includegraphics[width=\textwidth]{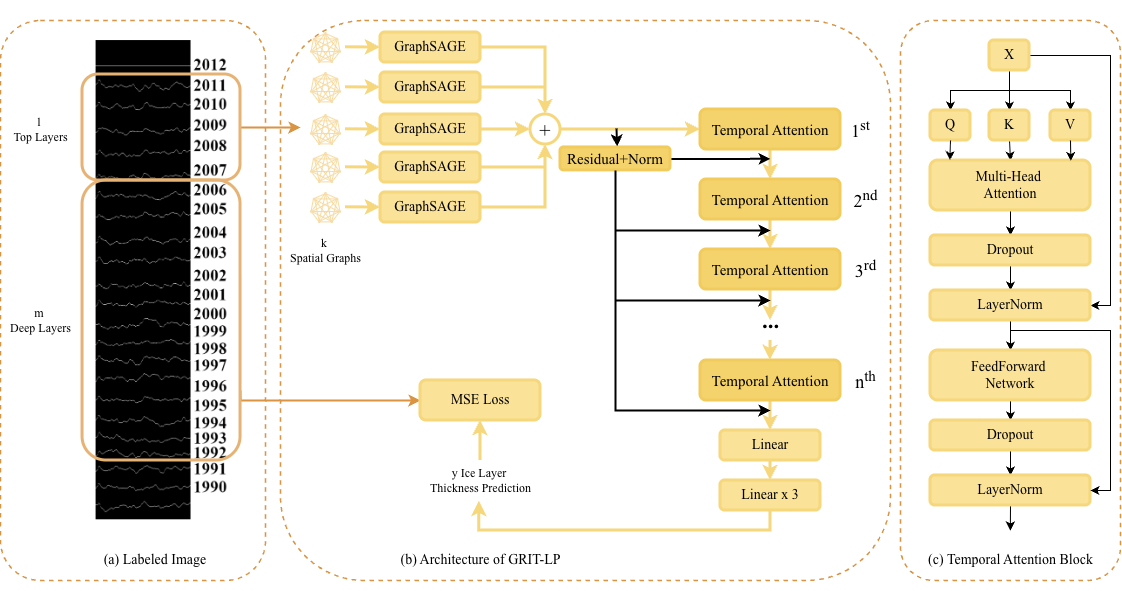}}
\caption{Network architecture of the proposed graph transformer network, GRIT-LP. (a) Overview of the complete network architecture.  (b) Architecture of GRIT-LP (c) Details of the temporal attention block. }
\label{fig:arch}
\end{center}
\end{figure*}

In this work, we introduce GRIT-LP, Graph Transformer with Long-Range Skip Connection and Partitioned Spatial Graphs for Accurate Ice Layer Thickness Prediction. It is a novel spatio-temporal graph transformer network designed to model the complex relationships between shallow and deeper ice layers. GRIT-LP utilizes the geographical and thickness information of the top $l$ ice layers to predict the thickness of the $m$ layers beneath (Figure \ref{fig:arch}), facilitating accurate predictions of deeper ice layer thickness. 

Built on a combination of self-attention backbone and geometric learning framework, GRIT-LP captures both local and long-range temporal relationships, effectively overcoming the restricted receptive field of traditional graph neural networks. It dynamically adjusts attention weights to prioritize critical input elements, reducing sensitivity to noise and irrelevant information. Moreover, the use of adaptive long-range skip connections balances raw spatial feature embeddings and learned temporal features after each attention block, dynamically mitigates key challenges in graph neural network like overfitting and over-smoothing. This approach facilitates the construction of deeper attention layers while preserving model performance, enabling the network to capture richer features and maintain stability in deeper architectures. In this section, we will introduce the building blocks of GRIT-LP, including GraphSAGE inductive geometric learning framework (Section~\ref{graphsage}), multi-head self-attention mechanism (Section~\ref{multi-head}), and our proposed adaptive long-range skip connection (Section~\ref{LR1}).

As shown in Figure \ref{fig:arch}, GRIT-LP takes temporal sequence of $k$ spatial graphs as inputs, and starts with a geometric deep learning part that composed of GraphSAGE\cite{hamilton2018inductive_graphsage} blocks. These blocks is designed to extract spatial patterns within each individual graphs and concatenate them together as spatial feature embeddings. This feature embedding is then passed into a sequence of $N$ temporal attention blocks. Adaptive long-range skip connections are used to connect the feature embedding to the output of each temporal attention block, which addresses key challenges such as overfitting and oversmoothing during the training process, enabling more robust and effective learning. Finally, a single linear layer is used to project the temporal dimension, and three linear layers with the hardswish activation function are applied to project the spatial dimension for the final prediction of the $m$ layers' thickness. 

\subsection{GraphSAGE Inductive Geometric Learning Framework}
\label{graphsage}
To capture spatial patterns, GRIT-LP utilizes GraphSAGE \cite{hamilton2018inductive_graphsage} layers to extract features from each independent ice layer and concatenate together to produce spatial feature embeddings. Unlike graph convolution network\cite{kipf2017_GCN}, it is an inductive framework designed to generate feature embeddings for unseen node via localized sampling and aggregation \cite{ZHOU202057_Review}. For an unseen node $i$, its node embedding is defined in Equation \ref{equ:graphsage}:
\begin{equation}
\textbf{x}'_i = \textbf{W}_1 \textbf{x}_i + \textbf{W}_2 \cdot \textit{mean}_{j \in \mathcal{N}(i)} \textbf{x}_j
\label{equ:graphsage}
\end{equation}
where $\textbf{x}_i$ is the feature matrix for the unseen node $i$, $\textbf{W}_1$ and $\textbf{W}_2$ are learnable weights of GraphSAGE, $\mathcal{N}(i)$ is the neighbor list of $i$, $\textbf{x}_j$ is the node feature matrix for these neighbor nodes and $\textit{mean}$ is the aggregation function. More specifically, GraphSAGE with the $\textit{mean}$ aggregation function shares a strong connection with GCNs, as it can be viewed as a linear approximation of localized spectral convolutions \cite{hamilton2018inductive_graphsage}. 

Compared with GCN, GraphSAGE has a few advantages. Graph Convolutional Networks rely on spectral formulations (like the graph Laplacian) and assume a fixed global graph structure, making them less suitable for our application where ice layer graphs vary across years and locations. In contrast, GraphSAGE is a spatial method that aggregates neighbor features using simple functions like mean, enabling it to operate directly on arbitrary and dynamic graph structures without depending on a global graph Laplacian. This flexibility allows GraphSAGE to generalize better across different and dynamic graph topologies, which aligns well with our need to model diverse ice layer structures.

Moreover, GraphSAGE separates the transformations for root and neighbor nodes, as shown in Equation\ref{equ:graphsage}, using distinct weight matrices $\textbf{W}_1$ and $\textbf{W}_2$. Here, $\textbf{W}_1$ operates on the root node feature, while $\textbf{W}_2$ applies to the aggregated neighbor features. This separation allows the model to learn how a node is influenced by its neighbors independently from how it retains its own identity. We interpret the term $\textbf{W}_1 \textbf{x}_i$ as functionally similar to a residual connection, enabling each node to preserve its unique geophysical characteristics rather than being entirely updated by surrounding context. This is particularly important for ice layer thickness prediction, where retaining node-specific information is crucial. Additionally, this design helps mitigate early over-smoothing before the attention-based encoders and supports better generalization to unseen data.

\subsection{Multi-Head Self-Attention}
\label{multi-head}
In the temporal encoder of our proposed GRIT-LP network, we employed a series of temporal attention blocks based on the standard multi-head self-attention mechanism proposed by \citet{vaswani2017attention}, as shown in Figure \ref{fig:arch}(\textbf{b}). By stacking multiple scaled dot-product attention applied on the temporal dimension, the temporal attention blocks can effectively capture the temporal dependencies in different range and enhance the overall model performance. For input feature embedding matrix $X$, the corresponding queries, keys, and values can be generated as:
\begin{equation}
    Q_i=XW_i^Q, K_i=XW_i^K, V_i=XW_i^V
\end{equation}
where $W_i^Q, W_i^K, W_i^V$ are both learnable weights matrix. In each head, self-attention score is then computed as: 
\begin{equation}
AttnScore(Q_i,K_i,V_i)=softmax(\frac{Q_iK_i^T}{\sqrt{d_k}})V_i
\end{equation}
where $Q_i, K_i, V_i$ are the calculated queries, keys and values, $d_k$ is the dimension of the key vector. In the end, results from each head are combined together as:
\begin{equation}
    \text{MHA}(X) = Concat(Head_1, Head_2,...,Head_n)W^o
\end{equation}

where $\text{MHA}$ stands for multi-head attention, $ Head_i = AttnScore(Q_i, K_i, V_i)$ and $W^o$ is a learnable matrix to concatenate all the results. As shown in Figure \ref{fig:arch}(\textbf{c}), in additional to the multi-head attention layer in the temporal attention block, those temporal attention block also contains feedforward network, layer normalization, dropout, and skip connection. In GRIT-LP we use 8 heads in total. Considering the fact that our task is a spatio-temporal learning task and our feature embeddings contains both a spatial dimension and a temporal dimension, necessary transpose operations are performed to apply the attention mechanism correctly on the temporal dimension.

\subsection{Adaptive Long-Range Skip Connection}
\label{LR1}
One of the key innovations in GRIT-LP is the introduction of adaptive long-range skip connections. Prior works such as DenseNet \cite{huang2018denselyconnectedconvolutionalnetworks} and U-Net \cite{ronneberger2015unetconvolutionalnetworksbiomedical} have shown that incorporating shortcut connection between early input and late layer outputs in convolutional architectures can enhance the accuracy and enable the training of deeper networks. These long-range connections have proven especially effective in dense prediction tasks \cite{DensePrediction}.

Motivated by these insights, we proposed a novel adaptive long-range skip connection mechanism tailored for graph transformers. Unlike conventional skip connections that locally link the input and output of each temporal attention block, our adaptive long-range skip connections directly bridge the raw spatial feature embeddings to the output of each temporal attention block, as illustrated in Figure\ref{fig:arch}. For the $i^{th}$ temporal attention block, we define its input as $\textbf{X}_{in}^i$ and output as $\textbf{X}_{out}^i$, where $i=1,2,...,n$, and $\textbf{X}_{in}^1=\textbf{X}$ is the initial feature embeddings that is the input of the first temporal attention block. The adaptive long-range skip connection can then be defined as:
\begin{equation}
\label{LR}
    \textbf{X}_{in}^i = \text{LayerNorm}((1-\alpha)\textbf{X}_{out}^{i-1} + \alpha \textbf{X})
\end{equation}
where $\alpha\in[0,1]$ is a learnable scalar parameter that can be initialized with different value. This design allows the model to dynamically interpolate between the updated temporal features and the original feature embeddings, providing a stable information anchor across layers. The inclusion of layer normalization ensures scale alignment between the two sources. This adaptive mechanism improves feature stability and mitigates the risks of gradient explosion when stacking multiple temporal attention layers. Moreover, it reduces prediction variability across input sequences, which is crucial for generalizing GRIT-LP to radargrams acquired in different years, by different sensors, or in varying regions of the ice sheet.

%% file: dataset.tex
\section{Generating Graphs From Radargram Dataset}
\label{data}

\begin{figure}[ht]
\begin{center}
\centerline{\includegraphics[width=\textwidth]{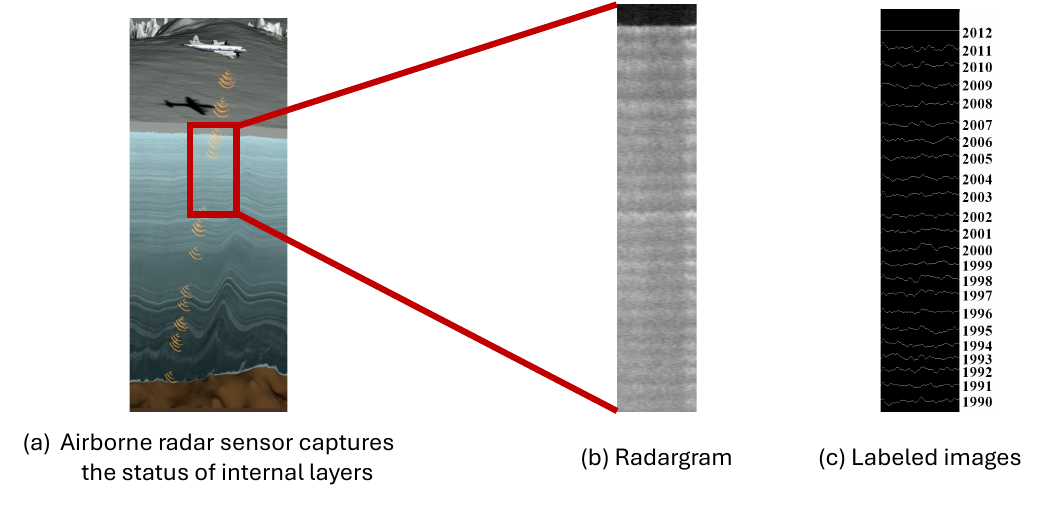}}
\caption{(\textbf{a}) Airborne radar sensor captures the status of internal ice layers by measuring the reflected signal.(Image adapted from \cite{diagram-airborne-radar}) (\textbf{b}) Radargram image (\textbf{c}) Labeled image, where boundaries of each ice layer is manually labeled out.}
\label{fig:dataset}
\end{center}
\end{figure} 

\subsection{Radargram Dataset}
We evaluate our proposed GRIT-LP on Snow Radar Echogram Dataset\cite{ibikunle2025aireadysnowradarechogram}, a radargram dataset contains about 14000 imagery of the Greenland region captured in 2012. This dataset was collected using an airborne snow radar sensor operated by CReSIS as part of NASA's Operation IceBridge \cite{Leuschen2011SnowRadar}. Each radargram has a fixed 256 pixels in width, with its depth varying from 1200 pixels to 1700 pixels. The dataset can be accessed at \url{https://data.cresis.ku.edu/data/snow/}.

Currently, airborne radar sensors have proven to be one of the most effective way to study internal ice layers, as radar sensor can penetrate the thick ice sheets and provide continuous measurements of internal ice layers over vast spatial areas. By measuring the strength of reflected signal(Figure \ref{fig:dataset} (\textbf{a})), airborne snow radar sensor capture the various depths of internal ice layer as radargrams (Figure \ref{fig:dataset} (\textbf{b})), where brighter pixels represents stronger reflection signal \cite{Arnold_2020}. Radargrams are then annotated to delineate the boundaries of each annual ice layer, producing the labeled images shown in Figure \ref{fig:dataset} (\textbf{c}). Using these annotated images, the thickness of each ice layer is determined by calculating the difference in coordinate values of its upper and lower boundaries. During radargram data collection, supplementary onboard systems are employed to record the latitude and longitude simultaneously.

\subsection{Graph Generation}
Graph dataset is generated by converting the tracked top $l$ ice layers of each radargram into a temporal sequence of $k$ spatial graphs, where each individual spatial graph represents a single ice layer in the top region of the radargram ($k=l$). Each spatial graph is composed of 256 nodes. Here, unlike current state-of-the-art methods that uses a fully connected graph, we proposed a novel localized connectivity on partitioned spatial graphs, as describe below. Edge weights are calculated based on the geographic distance between nodes, defined in Equation \ref{equ:edgeweight}:

\begin{equation}
        w_{i,j} = \frac{1}{2\arcsin{\big(hav(\phi_j-\phi_i) + \cos{\phi_i}\cos{\phi_j} hav(\lambda_j-\lambda_i)\big)}}
    \label{equ:edgeweight}
\end{equation}

where $i, j$ is any node in the spatial graph, $w_{i,j}$ is the edge weights, $\phi, \lambda$ are the latitude and longitude coordinates, and $hav(\theta) = \sin^2{(\frac{\theta}{2})}$. Each node will have three node features, which are latitude, longitude, and thickness.

\subsection{Localized Connectivity on Partitioned Spatial Graphs}

In all the baseline and current state-of-the-art models \cite{liu2024multibranchspatiotemporalgraphneural, zalatan_icip, Zalatan_igarss, Zalatan2023, liu2025_GRIT,liu2025_STGRIT}, each ice layer is modeled as a fully connected spatial graph, where every node is linked to all others regardless of their spatial proximity. Such global connectivity can obscure meaningful local structure and introduce unnecessary noise, ultimately hindering spatial pattern learning. To address this, we propose a novel localized connectivity scheme based on partitioned spatial graphs. Specifically, we divide each ice layer into fixed-size, fully connected local groups using a sliding window with partial overlaps. Within each group, all nodes are fully connected, forming a sequence of densely connected partitioned spatial graphs. There are no connections between nodes in different groups. The size of the sliding window can be any arbitrary number while in this work we use sliding window with size 5.

This partitioned design preserves strong local interactions critical for capturing spatial coherence while maintaining structural sparsity to avoid redundancy and irrelevant noisy long-range connections. By emphasizing spatially meaningful neighborhoods, our approach enables the model to learn cleaner and more robust spatial representations. This localized connectivity design also complements temporal modeling components by providing a clean and expressive spatial foundation on which to build spatio-temporal reasoning for ice layer thickness estimation.

%% file: experiments.tex
\section{Experiment Details}

\subsection{Data Preprocessing}
To highlight the performance of GRIT-LP, we compare it with the current state-of-the-art method ST-GRIT \cite{liu2025_STGRIT,liu2024multibranchspatiotemporalgraphneural}, as well as other baseline graph models including AGCN-LSTM \cite{zalatan_icip}, GCN-LSTM \cite{Zalatan2023}, GraphSAGE-LSTM \cite{liu2024learningspatiotemporalpatternspolar}, and GRIT \cite{liu2025_GRIT}.
We focus on a representative setting where $l=5$ and $m=15$, i.e., using the geographical and thickness information of the top 5 ice layers (formed during 2007–2011) to predict the thickness of the 15 underlying layers (formed during 1992–2006). 

Considering possible variations in snow accumulation, different melting process and different ice sheet topography, number of internal ice layers may varies at different location, resulting different number of ice layers in radargrams. To maintain overall high quality of our dataset, we perform a data preprocessing step by eliminating radargram images with fewer than 20 complete ice layers in the groundtruth. Image may be eliminated for insufficient number of ice layer available or the incompleteness of the top 20 layers due to melting or other physical process. After preprocessing, there are 1660 high quality radargrams with at least 20 complete ice layers, and we divided them into training, validation and testing sets with a ratio of $3:1:1$.

\subsection{Training Details}
\label{experiment-details}
All these networks were trained on the same machine with 8 NVIDIA A5000 GPUs and 2 Intel Xeon Gold 6430 CPUs. Mean-squared error (MSE) is used as the loss function for all the networks. Adam optimizer with 0.0001 as weight decay coefficient is used as the optimizer for all the networks. For those graph neural networks that don't contains attention blocks, we set the initial learning rate to be 0.01 and used a step learning rate scheduler that halves the learning rate every 75 epochs. We set the initial learning rate to be 0.001 for GRIT, 0.0005 for ST-GRIT and 0.0003 for our proposed GRIT-LP. Additionally, we employed an adaptive learning rate scheduler for the graph transformer networks that dynamically adjusts the learning rate based on validation loss. Specifically, the scheduler reduces the learning rate by half if performance stagnates for $s$ consecutive epochs without improvement, where $s=24$ for GRIT, $s=24$ for ST-GRIT, and $s=12$ for GRIT-LP. To ensure the fully convergence of each network, we trained all the networks for 450 epochs.

Both GraphSAGE and the adaptive learning rate scheduler may introduce some randomness to the training process and cause possible variation in the model's performance. In order to reduce the impact, we created five different version of training, validation, and testing dataset by applying random permutations to the whole dataset before splitting. Each network was trained on all these five versions and the average performance is reported as the model performance.

%% file: results.tex
\section{Results and Discussion}

\begin{table}[t]
\caption{Experiment results of AGCN-LSTM, GCN-LSTM, GraphSAGE-LSTM, Multi-Branch Spatio-Temporal GNN, GRIT, ST-GRIT, and our proposed GRIT-LP.}
\label{table:OverallResults}
\begin{center}
\scalebox{0.8}{
\begin{tabular}{ccc}
\toprule
Model & Average RMSE & Computation Time(Seconds)\\
\midrule
GCN \cite{kipf2017_GCN} & 5.0876 $\pm$ 0.1945 & 1098 \\
GraphSAGE\cite{hamilton2018inductive_graphsage} & 3.1383 $\pm$ 0.0550 & 834\\
AGCN-LSTM \cite{zalatan_icip}    & 3.4808 $\pm$ 0.0397 & 9404 \\
GCN-LSTM \cite{Zalatan2023} & 3.1745 $\pm$ 0.1045 & 7441\\
GraphSAGE-LSTM \cite{liu2024learningspatiotemporalpatternspolar}    & 3.3837 $\pm$ 0.1102 & 4579 \\
Multi-branch \cite{liu2024multibranchspatiotemporalgraphneural}    & 3.1087 $\pm$ 0.0555 & 987\\
\midrule
GRIT\cite{liu2025_GRIT} & 3.0597 $\pm$ 0.0326 & 1649\\
ST-GRIT\cite{liu2025_STGRIT} & 2.8866 $\pm$ 0.0569 & 2459\\
GRIT-LP($N=1$, $\alpha=0.5$ Ours) & 2.8907 $\pm$ 0.0703 & 1428\\
GRIT-LP($N=1$, $\alpha=0.75$ Ours) & 2.9148 $\pm$ 0.0801 & 1441\\
GRIT-LP($N=1$, $\alpha=0.25$ Ours) & 2.8976 $\pm$ 0.0597 & 1409\\
GRIT-LP($N=8$, $\alpha=0.5$, Ours) & 2.2195 $\pm$ 0.0660 & 3321\\
GRIT-LP($N=8$, $\alpha=0.75$, Ours) & 2.4621 $\pm$ 0.0446 & 3273\\
GRIT-LP($N=8$, $\alpha=0.25$, Ours, Best) &\textbf{2.1672 $\pm$ 0.0742} & 3368\\
\bottomrule
\end{tabular}
}
\end{center}
\end{table}

\begin{figure}[ht]
\begin{center}
\centerline{\includegraphics[width=0.9\textwidth]{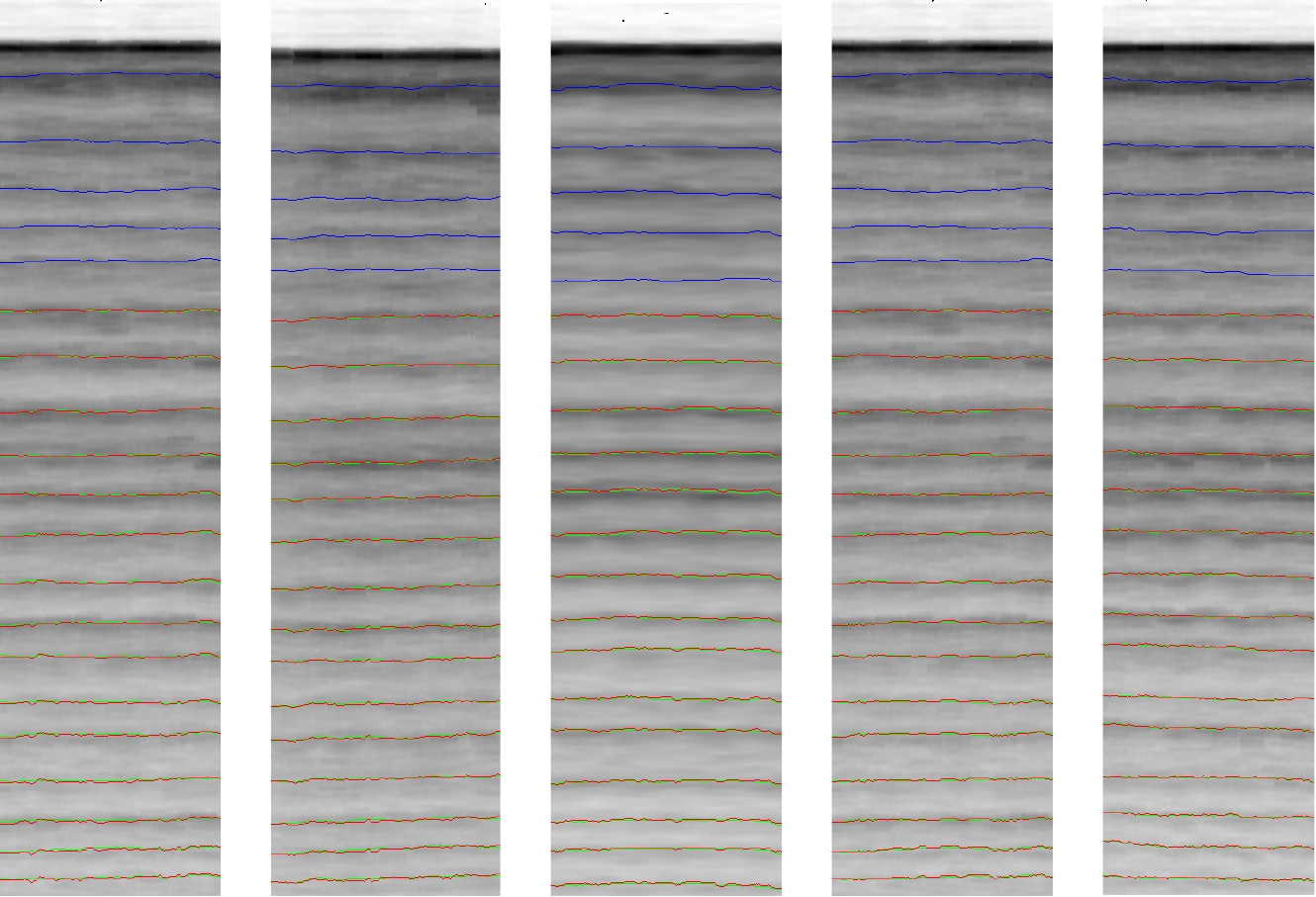}}
\caption{Qualitative visualization of some prediction results by GRIT-LP with $N=8$ and $\alpha=0.25$. The blue line is used to generate the graphs. The green line is the groundtruth (manually-labeled ice layers) and the red line is the model prediction.}
\label{fig:qual_GRIT}
\end{center}
\end{figure}

\subsection{Overall Performance}
For each training session, we measured the root mean squared error (RMSE) between the predicted thickness and the ground truth for the 15 deeper ice layers. Subsequently, for each approach, we computed the mean and standard deviation of the RMSE across five training runs on five different versions of the datasets. These metrics, representing the final performance of the models, are summarized in Table \ref{table:OverallResults}. Figure~\ref{fig:qual_GRIT} shows the qualitative prediction results of GRIT-LP with $N=8$ and $\alpha=0.25$.

Compared with current state-of-the-art model ST-GRIT\cite{liu2025_STGRIT}, our proposed GRIT-LP with one temporal attention block($N=1$) achieves a similar accuracy, and GRIT-LP with $N=8$ achieves the optimal RMSE of $2.1672 \pm 0.0742$, which is 24.92\% improvement compared with ST-GRIT. By removing explicit spatial attention blocks and introducing a low-cost adaptive long-range skip connection, GRIT-LP stacks more temporal attention blocks without increasing compute time. The skip connection dynamically balances raw spatial embeddings with temporally transformed features while preserving feature scale, yielding a richer, more stable representation that captures critical temporal relationships. More importantly, we also replace the previous fully connected spatial graphs with a novel localized connectivity strategy on partitioned spatial graphs. Instead of fully connecting all the nodes within each ice layers, we construct spatial graphs for each layer by dividing it into partially overlapping local neighborhoods, where nodes are fully connected only within each localized partition. This localized connectivity strategy enhances spatial coherence, suppresses noise from irrelevant long-range interactions, and enables the model to learn more robust and physically meaningful representations of ice layers. 

In the aspect of training time, we notice that compared with ST-GRIT, GRIT-LP with $N=1$ is about 41\% faster than ST-GRIT while maintains a similar performance in accuracy. GRIT-LP with $N=8$ is about 37\% slower than ST-GRIT while achieves a 24.92\% improvement in accuracy. Compared with the multi-branch spatio-temporal graph neural network that has the best efficiency, GRIT with $N=1$ is about 47\% slower and GRIT with $N=8$ is about 241\% slower. However, we should also notice that the multi-branch spatio-temporal graph neural network is specifically optimized for the best efficiency, while the proposed GRIT-LP is specifically designed for the best accuracy. In our application scenario, accuracy of ice layer thickness prediction is more important than efficiency as we want to lower the uncertainties of downstream tasks. Our experiments also confirm that GRIT-LP remains highly scalable. It requires less than 1GB GPU memory per GPU when training GRIT-LP with $N=8$ on 8 GPUs, and the SRED Dataset is the current largest dataset for internal ice layers. 

\begin{figure*}[ht]
\begin{center}
\centerline{\includegraphics[width=\textwidth]{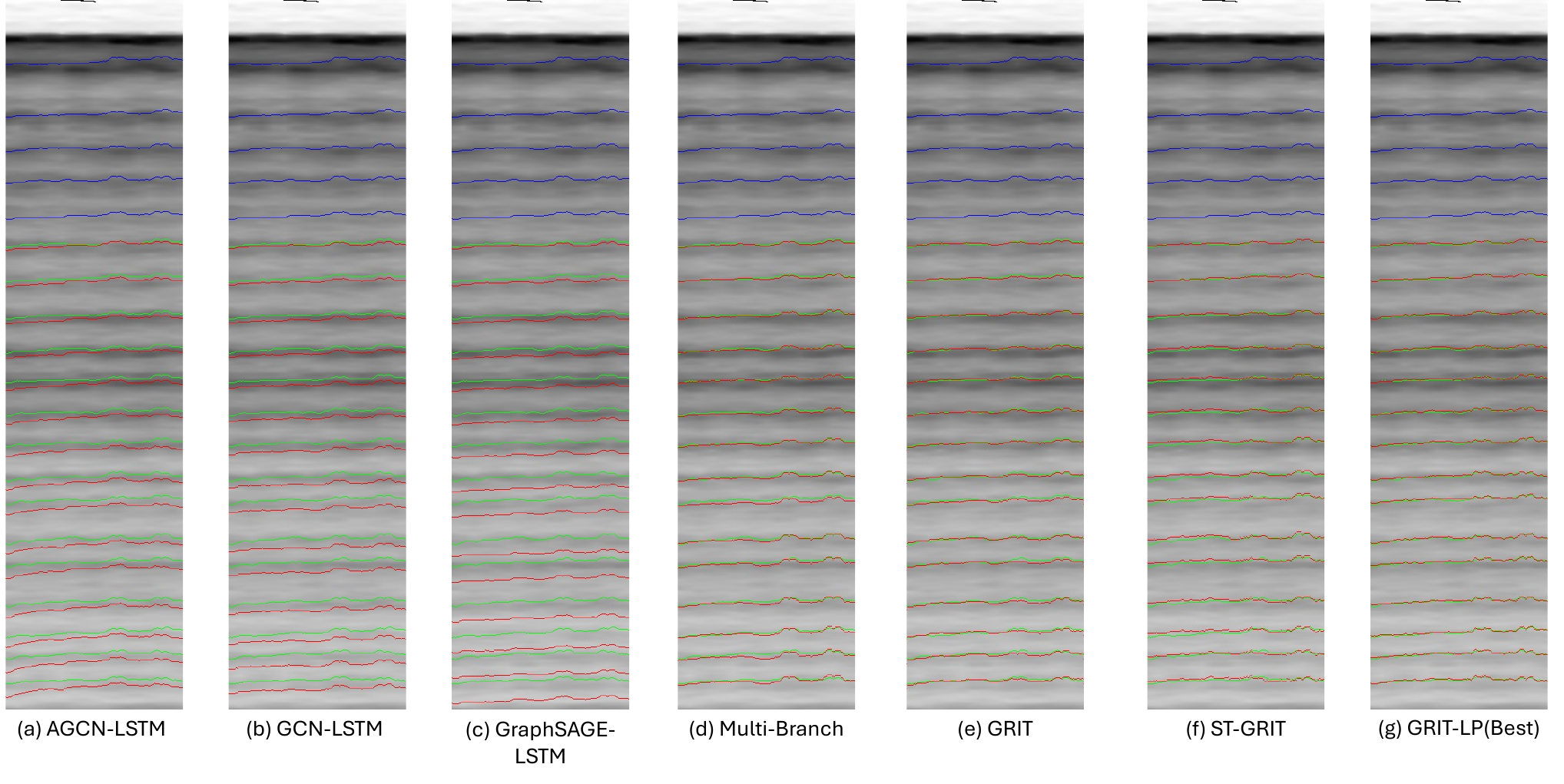}}
\caption{Qualitative results that shows a comparison of different model predictions on the same radargram. The blue line is used to generate the graphs. The green line is the groundtruth (manually-labeled ice layers) and the red line is the model prediction. "Best" means GRIT-LP with 8 attention blocks and $\alpha=0.25$.}
\label{fig:qualitative}
\end{center}
\end{figure*}

Figure \ref{fig:qualitative} shows a qualitative comparison results, where we compared model predictions on the same radargram. We notice that for this radargram, a shift between groundtruth and model prediction exists in the results of AGCN-LSTM, GCN-LSTM, and GraphSAGE-LSTM, which is a sign of error accumulation. Current state-or-the-art ST-GRIT\cite{liu2025_STGRIT} somehow address this issue, while GRIT-LP provides notable improvement in avoiding error accumulation towards those most deeper ice layer. Moreover, we also notice that compared with current state-of-the-art method and other baseline methods, our proposed GRIT-LP improves the prediction accuracy for the left and right boundary region of the radargram. This results show that GRIT-LP still maintains a decent performance for few-pixel-level localized patterns while improve the overall capacity in capture long-range temporal relationship. High boundary region prediction accuracy significantly impact downstream tasks like modeling glacier dynamics. We will discuss more about performance on boundary regions in later section.

\subsection{Discussion on the choice of \texorpdfstring{$\alpha$}{alpha}}
In our proposed adaptive long-range skip connection, $\alpha$ in Equation~\ref{LR} is a hyperparameter that is used to balance how much of the original spatial feature embedding is retrained versus how much is replaced by the updated temporal features. This balance is crucial: a large $\alpha$ risks over-reliance on spatial features, while a small $\alpha$ may overemphasize temporal features, causing overfitting and unstable gradients. In our experiment, we evaluate the performance of GRIT-LP with $N=1$ and $N=8$ on three different initial value of $\alpha$: 0.25, 0.5, 0.75. Results are shown in Table~\ref{table:alpha}

\begin{table}[t]
\caption{Experiments of GRIT-LP with $N=1$ and $N=8$ on $\alpha$ with different values. RMSE of each individual trial is reported together with the final average RMSE}
\label{table:alpha}
\begin{center}
    \scalebox{0.8}{
    \begin{tabular}{cccccccc}
    \toprule
N & Alpha & Trial 1 & Trial 2 & Trial 3 & Trial 4 & Trial 5 & Average RMSE      \\ \midrule
1 & 0.25  & 2.8152 & 2.9201 & 2.9501 & 2.9636 & 2.8393 & 2.8976 $\pm$ 0.0597   \\
1 & 0.50  & 2.8272 & 2.8510 & 3.0235 & 2.8983 & 2.8535 & \textbf{2.8907 $\pm$ 0.0703}  \\
1 & 0.75  & 2.7846 & 2.9220 & 3.0335 & 2.8942 & 2.9398 & 2.9148 $\pm$ 0.0801 \\ \midrule
8 & 0.25  & 2.1417 & 2.2057 & 2.2926 & 2.1039 & 2.0921 & \textbf{2.1672 $\pm$ 0.0742}  \\
8 & 0.50  & 2.1568 & 2.2438 & 2.3370 & 2.1676 & 2.1922 & 2.2195 $\pm$ 0.0660 \\
8 & 0.75  & 2.4162 & 2.4678 & 2.5452 & 2.4382 & 2.4434 & 2.4621 $\pm$ 0.0446 \\ \bottomrule
\end{tabular}
    }
\end{center}
\end{table}

Our experiments show that for both $N=1$ and $N=8$, $\alpha=0.75$ slightly degrades performance by overly favoring residual features. $\alpha=0.25$ achieves a lower RMSE error on each individual trial with $N=8$ and thereby a notable lower average RMSE in the end, while having a mixing performance on different trials with $N=1$. These results suggests that there is no single rule for choosing the optimal $\alpha$ value. It is sensitive to both number of attention layers used in the model, and characteristics that related to the dataset itself.

\subsection{Ablation Study}
There are four major components in our proposed method: GraphSAGE inductive learning framework, attention-based temporal blocks, novel adaptive long-range skip connection and novel localized connectivity strategy. In order to assess the contribution of each components, we conduct a detailed ablation study. Table \ref{table:ablation} shows results for all meaningful combinations of these four components.

\begin{table}[t]
\caption{Ablation study results on all the meaningful combinations of four major components with $N=8$ and $\alpha=0.25$. "Graph" means using 5 GraphSAGE layer. "Attention" means using 8 temporal attention blocks. "Adaptive LR Connection" means using our proposed adaptive long-range skip connection. "Localized Connectivity" means using our proposed localized connectivity strategy based on partitioned spatial graphs.}
\label{table:ablation}
\begin{center}
\scalebox{0.7}{
\begin{tabular}{ccccc}
\toprule
Graph & Attention & Adaptive LR Connection ($\alpha$=0.25) & Localized Connectivity & RMSE\\
\midrule
\checkmark & x & x & x & 3.4443 $\pm$ 0.0361\\
\checkmark & x & x & \checkmark & 3.4573 $\pm$ 0.0540\\
\midrule
x & \checkmark & x & x & 5.7035 $\pm$ 0.7874\\
x & \checkmark & x & \checkmark & 5.7035 $\pm$ 0.7874\\
x & \checkmark & \checkmark & x & 5.1185 $\pm$ 0.1946\\
x & \checkmark & \checkmark & \checkmark & 5.1185 $\pm$ 0.1946\\
\midrule
\checkmark & \checkmark & x & x & 2.6637 $\pm$ 0.1385\\
\checkmark & \checkmark & \checkmark & x & 2.5487 $\pm$ 0.0991\\
\checkmark & \checkmark & x & \checkmark & 2.2207 $\pm$ 0.1378\\
\checkmark & \checkmark & \checkmark & \checkmark & \textbf{2.1672 $\pm$ 0.0742}\\
\bottomrule
\end{tabular}
}
\end{center}
\end{table}

We group the ablation results into three categories. As shown in the table, using only 5 GraphSAGE layers or only 8 temporal attention blocks results in a higher RMSE, indicating that our task requires spatio-temporal pattern learning. Both the spatial and temporal components of the network are essential for accurate prediction. In studying the effect of localized graph connectivity, we observe that applying localized graph connectivity can bring about 16\% improvement in accuracy. This finding suggests that fully connected spatial graphs may introduce redundant interactions between distant nodes, making it harder for the model to focus on critical local structures. In contrast, our localized connectivity promotes structural sparsity, enabling the model to concentrate on spatially coherent and meaningful relationships, thereby enhancing learning efficiency and predictive accuracy.

In studying the effect of our proposed adaptive long-range skip connection with $\alpha=0.25$, we found a 3\%-5\% improvement in average RMSE and a modest improvement in its standard deviation. These results indicate that without the skip connection, the model begins to forget learned spatial features after deep temporal attention blocks, suffers from poor generalization ability, potentially leading to large errors when applied to unseen radargrams captured in different years, by different radar sensors, or across various locations of the ice sheet. By introducing an adaptive long-range skip connection that links the learned spatial embeddings to the output of each temporal attention block, the model receives a fresh infusion of spatial features at every depth while preserving feature scale, mitigating feature drift and over-smoothing. This yields a richer, more stable representation, enabling deeper temporal stacks with reduced error and variance and improved robustness to distribution shifts.

Therefore, our experiments prove that all the 4 components of GRIT-LP are indispensable. Each contributes uniquely to the model’s ability to learn complex spatio-temporal patterns from internal ice layer data, enable GRIT-LP to achieve both high accuracy and robustness, making it well-suited for reliable ice layer thickness prediction in real-world scenarios.

\subsection{Performance on Boundary Region of Radargrams}
\label{boundary_pixel}
Accurately tracking the boundary regions of ice layers in radargram blocks is essential for merging them into a seamless spatiotemporal assessment of snow accumulation variability over an extremely large spatial area. The precision of this reconstructed spatiotemporal variation plays a critical role in downstream tasks, such as monitoring glacial dynamics, ultimately contributing to a better understanding of climate change. Here, we proposed a $p\text{-pixel}$ boundary RMSE defined in Equation \ref{equation_pixel}:

\begin{equation}
    \textit{RMSE}_{p\text{-pixel}} = 
    \sqrt{\frac{1}{m \times 2p} \sum_{i=1}^{m} \sum_{j=1}^p 
    \big[(y_{i,j} - \hat{y}_{i,j})^2 
    {+ (y_{i, w-p+j} - \hat{y}_{i, w-p+j})^2\big]}}
\label{equation_pixel}
\end{equation}

where $m$ is the number of deep ice layer that we are predicting the thickness, $p$ is the number of boundary pixels extracted from the left and right of each layer, and $w$ is the width of the radargrams ($w=256$ for this work). This metrics aims to quantify model performance on the left and right boundary regions of radargrams. We calculated the 1-pixel, 2-pixel, 5-pixel, and 10-pixel error for GRIT-LP with the optimal number of attention blocks, and compare it with other methods. Similarly, we also calculate the mean and standard deviation of the $p\text{-pixel}$ boundary RMSE.

\begin{table}[h]
\caption{Experiment Results for Boundary Pixels of Each Radargram}
\label{table:boundary}
\begin{center}
\scalebox{0.63}{
\begin{tabular}{lccccc}
\toprule
                   Methods                   & 1-Pixel& 2-Pixel& 5-Pixel & 10-Pixel\\ \midrule
AGCN-LSTM \cite{zalatan_icip}              & 3.9106 $\pm$ 0.3356 & 3.7401 $\pm$ 0.1970 & 3.6262 $\pm$ 0.1066 & 3.5870 $\pm$ 0.0718\\
GCN-LSTM \cite{Zalatan2023}                & 3.5436 $\pm$ 0.3702 & 3.3570 $\pm$ 0.2298 & 3.2370 $\pm$ 0.1405 & 3.2018 $\pm$ 0.1172\\
SAGE-LSTM \cite{liu2024learningspatiotemporalpatternspolar}                        & 3.7634 $\pm$ 0.2903 & 3.5840 $\pm$ 0.1596 & 3.4707 $\pm$ 0.0956 & 3.4353 $\pm$ 0.0965 \\
Multi-branch \cite{liu2024multibranchspatiotemporalgraphneural}                          & 3.5270 $\pm$ 0.3605 & 3.3365 $\pm$ 0.2080 & 3.2114 $\pm$ 0.1063 & 3.1739 $\pm$ 0.0760\\
GRIT\cite{liu2025_GRIT} & 3.6842 $\pm$ 0.2945 & 3.4057 $\pm$ 0.1658 & 3.2190 $\pm$ 0.0781 & 3.1699 $\pm$ 0.0580 \\
ST-GRIT\cite{liu2025_STGRIT} & 3.3880 $\pm$ 0.3074 & 3.1876 $\pm$ 0.1515 & 3.0507 $\pm$ 0.0645 & 3.0194 $\pm$ 0.0581 \\
GRIT-LP(8 Blocks, $\alpha=0.25$, Ours) & \textbf{2.7379 $\pm$ 0.3921} & \textbf{2.5089 $\pm$ 0.2418} & \textbf{2.3063 $\pm$ 0.1652} & \textbf{2.2362 $\pm$ 0.1432}\\ \bottomrule
\end{tabular}
}
\end{center}
\end{table}  

From Table \ref{table:boundary}, we can see that compared to baseline GNNs and current state-of-the-art multi-branch spatio-temporal graph neural network\cite{liu2024multibranchspatiotemporalgraphneural}, our proposed GRIT-LP demonstrates a significant improvement in predicting the boundary pixels of each ice layer. This results consolidate the conclusion we got from Figure \ref{fig:qualitative}, underscore that besides the ability to effective capture the long-range temporal patterns, the use of temporal attention blocks can also have outstanding performance in capture small-scale patterns that are typically noisy and hard to learn.

%% file: conclusion.tex
\section{Conclusion}

In this paper, we introduce GRIT-LP, a graph transformer network enhanced with adaptive long-range skip connections and localized graph connectivity. GRIT-LP is specifically designed for ice layer thickness prediction, aiming to learn from the geographical and thickness information of shallow ice layers and make predictions for deeper layers. GRIT-LP employs the GraphSAGE inductive framework to extract spatial patterns within each individual graph and uses a few temporal attention blocks to capture the temporal variations in different scales. GRIT-LP introduces a novel adaptive long-range skip connection that reinjects spatial embeddings into every temporal block, with the hyperparameter $\alpha$ tuning the balance between raw spatial embeddings and transformed temporal features. This design preserves feature scale, mitigates over-smoothing, and enables deeper, more stable temporal stacks with stronger generalization. We further propose a novel localized connectivity strategy based on partitioned spatial graphs, which promotes structural sparsity by focusing on critical spatial structures while effectively avoiding irrelevant long-range connections.

We evaluate GRIT-LP on a real-world application dataset, using data from internal ice layers of the Greenland region formed between 2007 and 2011 to predict the thickness of layers formed from 1992 to 2006. Notably, GRIT-LP can be applied to predict the thickness of varying numbers of ice layers and radagrams of different sizes and different locations. Experiments show that GRIT-LP achieves the best performance with 8 temporal attention blocks and $\alpha=0.25$, resulting in approximately 24.92\% improvement compared to the current state-of-the-art model. Moreover, GRIT-LP always maintains a low standard deviation, demonstrating its robustness and reliability across varying radargrams, including those from different locations and time periods. This strong generalization ability makes it a valuable asset for downstream applications, such as understanding glacial flow dynamics and contributing to climate change research.